%% file: NeuVideoRec.tex
\documentclass[10pt,twocolumn,letterpaper]{article}

\usepackage[pagenumbers]{cvpr} 

\usepackage{graphicx}
\usepackage{amsmath}
\usepackage{amssymb}
\usepackage{booktabs}
\usepackage{multirow}

\usepackage[pagebackref,breaklinks,colorlinks]{hyperref}

\usepackage[capitalize]{cleveref}
\crefname{section}{Sec.}{Secs.}
\Crefname{section}{Section}{Sections}
\Crefname{table}{Table}{Tables}
\crefname{table}{Tab.}{Tabs.}

\def\cvprPaperID{8135} 
\def\confName{CVPR}
\def\confYear{2022}

\begin{document}




\title{SelfRecon: Self Reconstruction Your Digital Avatar from Monocular Video}
\author{\large Boyi Jiang\textsuperscript{1,2}\quad Yang Hong\textsuperscript{1} \quad Hujun Bao\textsuperscript{3} \quad Juyong Zhang\textsuperscript{1}\thanks{Corresponding author} \vspace{1.5 mm}\\
	{\normalsize \textsuperscript{1}University of Science and Technology of China \quad
		\textsuperscript{2}Image Derivative Inc \quad \textsuperscript{3}Zhejiang University}\\
}
\maketitle

\begin{abstract}
We propose SelfRecon, a clothed human body reconstruction method that combines implicit and explicit representations to recover space-time coherent geometries from a monocular self-rotating human video. Explicit methods require a predefined template mesh for a given sequence, while the template is hard to acquire for a specific subject. Meanwhile, the fixed topology limits the reconstruction accuracy and clothing types. Implicit representation supports arbitrary topology and can represent high-fidelity geometry shapes due to its continuous nature. However, it is difficult to integrate multi-frame information to produce a consistent registration sequence for downstream applications. We propose to combine the advantages of both representations. We utilize differential mask loss of the explicit mesh to obtain the coherent overall shape, while the details on the implicit surface are refined with the differentiable neural rendering. Meanwhile, the explicit mesh is updated periodically to adjust its topology changes, and a consistency loss is designed to match both representations. Compared with existing methods, SelfRecon can produce high-fidelity surfaces for arbitrary clothed humans with self-supervised optimization. Extensive experimental results demonstrate its effectiveness on real captured monocular videos. The source code is available at \href{https://github.com/jby1993/SelfReconCode}{https://github.com/jby1993/SelfReconCode}.
\end{abstract}

\input{introduction.tex}

\input{related_work.tex}

\input{algorithm.tex}

\input{experiment.tex}

\input{conclusion.tex}

{\small
\bibliographystyle{ieee_fullname}
\bibliography{egbib}
}

\end{document}

%% file: introduction.tex
\section{Introduction}
Clothed body reconstruction has been an important and challenging research topic in the community for years. In the film and gaming industry, high-fidelity human reconstruction usually requires pre-captured templates, multi-camera systems, controlled studios, and long-term works of talented artists. However, these requirements exceed the application scenarios of general customers, such as personalized avatars for telepresence, AR/VR, anthropometry, and virtual try-on, etc. Therefore, directly reconstruction high-fidelity digital avatar from monocular video will have significant practical application value.

The state-of-the-art marker-less monocular human performance capture approaches~\cite{xu2018monoperfcap,habermann2019livecap,habermann2020deepcap} are mainly designed based on explicit mesh representation. They
require actor-specific rigged templates and utilize detected 2D/3D joints and silhouettes to estimate per frame's posture and non-rigid deformation. DeepCap~\cite{habermann2020deepcap} additionally uses multi-view information during training to resolve deep ambiguity and improve tracking accuracy for monocular inference. The explicit representation has some advantages, including space-time coherence and compatibility with existing graphics control pipelines, like texture editing and reposing. Moreover, skinning deformation is suitable under this paradigm to model the body's large-scale articulated deformations. However, actor-specific templates limit the extension of these methods to unseen human sequences. For videos of self-rotation humans under rough A-pose, VideoAvatar~\cite{alldieck2018video} can estimate general clothed humans with the SMPL+D parametric representation~\cite{loper2015smpl,bhatnagar2019mgn,alldieck2019tex2shape,alldieck19cvpr,alldieck2018video,xiang2020monoclothcap}, while it can not recover folds and loose clothing, like skirts.

Recently, some neural implicit representation based monocular human reconstruction approaches have demonstrated compelling results~\cite{zheng2019deephuman,saito2019pifu,saito2020pifuhd,zheng2021pamir,MetaAvatar:NeurIPS:2021,huang2020arch,he2021arch++,he2020geopifu,burov2021dsfn,xiu2021icon}. These methods can handle various topologies, and thus can represent various clothing and hairstyles. However, they require high-quality 3D data for supervision, and they only reconstruct for a specific frame and can not keep the space-time coherence of surface vertices for the whole sequence. A simple solution to guarantee the coherence and correct body structure is to maintain an implicit template surface in the canonical space, and then utilize backward deformation fields to map current points to canonical space to assist their implicit function queries. The backward deformation strategy has been widely applied recently and works well for small-scale deformations~\cite{park2021nerfies,pumarola2021d,tretschk2021non,li2021neural}. However, it is not very suitable for articulated skinning deformation due to its irreversibility in some parts of current space~\cite{jeruzalski2020nilbs,chen2021snarf}. To this end, technologies such as pose-related skinning weights prediction~\cite{weng2020vid2actor,peng2021animatable,jeruzalski2020nilbs} and specific inverse articulated deformation design~\cite{deng2020nasa} are proposed at the cost of high complexity and poor generalization.

In this work, we propose SelfRecon, which combines the explicit and implicit representations together to reconstruct high-fidelity digital avatar from a monocular video. Specifically, SelfRecon utilizes a learnable signed distance field (SDF) rather than a template with fixed topology to represent the canonical shape. To improve the generalization of the deformation and reduce the optimization difficulty, we adopt the forward deformation to map canonical points to the current frame space~\cite{chen2021snarf,zheng2021avatar}. During optimization, we periodically extract the explicit canonical mesh and warp it to each frame with the deformation fields. For these meshes, we utilize mask loss and smooth constraints to recover the overall shape. For the implicit part, a differential formulation is designed to intersect the deformed surface and follow IDR's neural rendering~\cite{yariv2020multiview} to refine the geometry. A consistency loss is designed to match both geometric representations as close as possible.

SelfRecon alleviates the dependence on actor-specific templates and extracts a space-time coherent mesh sequence from a monocular video. Extensive evaluations on self-rotating human videos demonstrate that it outperforms existing methods. We believe that SelfRecon will inspire more studies on combining implicit and explicit representations for 3D reconstruction for articulated object.




%% file: related_work.tex
\section{Related Work}
\textbf{Implicit Human Reconstruction.} PIFu~\cite{saito2019pifu} adopts a deep network to extract image features and concatenates pixel's feature and its corresponding 3D point depth information as the input of a Multi-Layer Perceptron (MLP) to obtain high-fidelity 3D clothed human occupancy field. However, it may generate incorrect body structures for humans under challenging poses. StereoPIFu~\cite{yang2021stereopifu} aims at binocular images, utilizes volume alignment feature and predicted high-precision depth to guide implicit function prediction, can effectively alleviate the depth ambiguity and restore absolute scale information. PIFuHD~\cite{saito2020pifuhd} utilizes higher resolution features and predicted normal information to refine the geometric details of PIFu. PaMIR~\cite{zheng2021pamir} utilizes parameterized human body to decrease the influence of deep ambiguity in implicit function training, reduces the occurrence of abnormal human body structure, and improves the reconstruction accuracy. These methods train an MLP to represent the human's implicit geometry from single or several images and achieve impressive results. However, they require the corresponding high-quality 3D data of color images to train the model, which is hard to obtain and thus limits their generalization to in-the-wild images.

Besides, overfitting the implicit neural representation of a person's movement sequence to acquire actor-specific reconstructions becomes popular. NASA~\cite{deng2020nasa} coarsely models the naked body as the union of articulated parts, and each part is an implicit occupancy field. SCANimate~\cite{saito2021scanimate} proposes an end-to-end trainable framework that turns raw 3D scans of a clothed human into an animatable avatar. SNARF~\cite{chen2021snarf} learns a forward deformation field to improve its generalization for unseen human poses. All these methods need 4D scan data to train their clothed body representation, and thus are difficult to be widely used for general image data.

Recently, some implicit representation methods, which can extract geometry and synthesize novel views based on multi-view images, attract researchers' attention. NeuralBody~\cite{peng2021neural} reconstructs per frame's NeRF~\cite{mildenhall2020nerf} field conditioned at body structured latent codes and utilizes the NeRF field to synthesize new images. However, the extracted geometry from NeRF suffers from noise. H-NeRF~\cite{xu2021h} utilizes an implicit parametric model~\cite{alldieck2021imghum} to reconstruct the temporal motion of humans. Neural Actor~\cite{liu2021neuralactor} integrates texture map features to refine volume rendering. IDR~\cite{yariv2020multiview} combines implicit signed distance field and differential neural rendering to generate high-quality rigid reconstruction from multi-view images. Concurrent IMAvatar~\cite{zheng2021avatar} expand IDR to learn implicit head avatars from monocular videos.

\textbf{Explicit Human Reconstruction.} With the help of human statistical model~\cite{anguelov2005scape,loper2015smpl,Jiang2020HumanBody}, some works utilize image cues to automatically obtain model parameters~\cite{bogo2016keep,guler2018densepose,kanazawa2018end,omran2018neural}. To represent human clothing, some methods add displacements on SMPL~\cite{loper2015smpl} vertices to model tight clothing~\cite{pavlakos2018learning,bhatnagar2019multi,ma2020learning,alldieck2018video}. However, this SMPL+D representation can only support tight clothing types and recover coarse level geometry shape. To improve the representation ability, some works adopt separate clothing representation and combine with SMPL body to do reconstruction~\cite{jiang2020bcnet,ma2021scale}, but they need clothing type and high-quality 3D supervision.

Besides, to capture the performance of a specific person, many prior works use an actor-specific template to assist tracking. Monoperfcap~\cite{xu2018monoperfcap} optimizes the deformation of template mesh to match 2D cues. LiveCap~\cite{habermann2019livecap} refines the optimization pipeline and achieves real-time tracking for a specific person with monocular RGB input. DeepCap~\cite{habermann2020deepcap} adopts a network to predict per frame's template deformation for a specific person. However, the requirement for pre-defined templates limits their broader applications.

%% file: algorithm.tex
\section{Method}
\begin{figure*}
	\begin{center}
		\includegraphics[width=\linewidth]{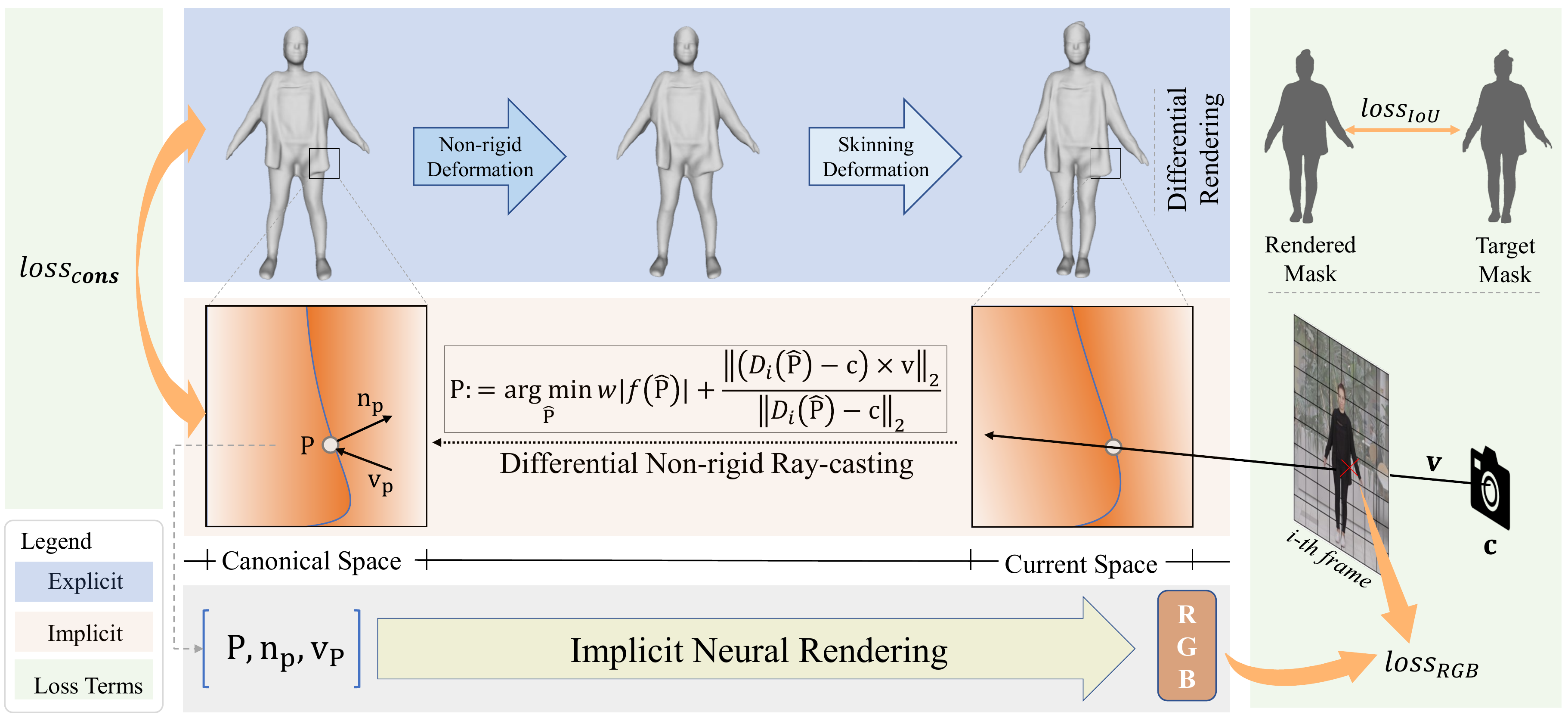}
	\end{center}
	\caption{The pipeline of SelfRecon. We simultaneously maintain the explicit and implicit geometry representations and use forward deformation fields to transform canonical geometry to the current frame space. For explicit representation, we mainly use differentiable mask loss to recover the overall shape. As for implicit representation, sampled neural rendering loss and predicted normals are used to refine geometry details. Finally, a consistency loss is used to keep both geometric representations matched.}
	\label{fig:pipeline}
\end{figure*}

SelfRecon aims to reconstruct a high-fidelity and space-time coherent clothed body shape from a monocular video depicting a self-rotating person, and the whole algorithm pipeline is given in Fig.~\ref{fig:pipeline}. Both explicit and implicit geometric representations are utilized to achieve the above target. Specifically, we utilize the forward deformation field to generate space-time coherent explicit meshes. The deformation fields are decomposed into two parts, where the first one represents per frame's non-rigid deformation with a learnable MLP, and the second is the skinning deformation field. Differentiable masks, regular and smooth losses are adopted to control the shape of explicit meshes. To update the shape of implicit neural representation, we use non-rigid ray-casting (sec~\ref{sec:nonrigid_raycast}) to find the differentiable intersection points of rays and the deformed implicit surface. Then, the implicit rendering network (sec~\ref{sec:implicit_render}) will utilize the rays' color information to improve the geometry. Unless otherwise indicated, we also utilize the predicted normal map~\cite{saito2020pifuhd} to refine the details. Finally, a consistency loss is designed to match both representations.

For a self-rotating video with $N$ frames, we adopt the method described in VideoAvatar~\cite{alldieck2018video} to generate the initial shape parameter $\boldsymbol{\beta}$ and per-frame's pose parameters $\{\boldsymbol{\theta}_i|i\in\{1,...,N\}\}$ of SMPL model. We pre-defined a template pose and generate initial canonical SMPL body mesh $\mathcal{B}$ with $\boldsymbol{\beta}$ and this pose parameter. Our implicit and explicit representations are both initialized with $\mathcal{B}$. In the following, we present the algorithm details of each component.

\subsection{Canonical Implicit SDF}
In the similar work of VideoAvatar~\cite{alldieck2018video}, they adopt the SMPL+D representation for clothed human body. However, SMPL+D has limited resolution and representation ability, and thus it can not represent high-fidelity geometry shape and various clothing types. In this work, we represent the canonical template shape $\mathcal{S}_{\eta}$ as the zero isosurface of a SDF, which is expressed by an MLP $f$ with learnable weights $\eta$:
\begin{equation}
\mathcal{S}_{\eta} = \{\mathbf{p}\in{\mathbb{R}^3}|f(\mathbf{p};\eta) = 0\}.
\end{equation}
To avoid unexpected solution, we use IGR~\cite{gropp2020implicit} to initialize $\mathcal{S}_{\eta}$ as the initial canonical body $\mathcal{B}$. 

\subsection{Deformation Fields}
\label{sec:deform}
Following prior works~\cite{huang2020arch,he2021arch++}, we utilize skeleton skinning transformation to control human body's large-scale movements due to the articulated structure. However, garments' non-rigid deformation cannot be fully represented by skinning transformation. Therefore, we extend to model non-rigid deformation with another MLP.

\textbf{Non-rigid Deformation Field.} We use an MLP $d$ with learnable weights $\phi$ to represent the non-rigid deformation field. For $i$-th frame, $d$ takes its optimizable conditional variable $\mathbf{h}_i$ as input and deform points in the canonical space with $i$-th frame's specific non-rigid deformation.

\textbf{Skinning Transformation Field.} Given $i$-th frame's pose parameter $\boldsymbol{\theta}_i$, we have to define a canonical-to-current space skinning transformation field $\mathcal{W}$. As the initial template body $\mathcal{B}$ has well-defined skinning weights relate to its SMPL skeleton, an intuitive idea is to expand the skinning weights of $\mathcal{B}$'s vertices to the whole canonical space to define the skinning transformation field. Specifically, we first pre-defined a sparse grid containing $\mathcal{B}$ in the canonical space. For each grid point, we find its nearest $30$ vertices on $\mathcal{B}$ and average their skinning weights with IDW (inverse distance weight) as its initial weight. Then, we smooth all grid points' weights with Laplace smoothing. Finally, given a point in the canonical space, we compute its skinning weights by trilinear interpolation in the grid. During our optimization, the grid is pre-computed and fixed. This forward deformation design avoids the trouble of inverse skinning transformation~\cite{weng2020vid2actor,deng2020nasa,jeruzalski2020nilbs,chen2021snarf} and provides a regular constrain for human articulated movement. 

Finally, by compositing $d$ and $\mathcal{W}$, we get the final deformation field $\mathcal{D} = \mathcal{W}(d(\cdot))$. It takes $i$-th frame's conditional variable $\mathbf{h}_i$ and SMPL pose parameter $\boldsymbol{\theta}_i$ as input, and transform canonical points to the $i$-th frame space. For brevity of description, we use $\mathcal{D}_i$ to denote $i$-th frame's deformation field, $\mathcal{S}_i$ for $i$-th frame's zero isosurface $\mathcal{D}_i(\mathcal{S}_{\eta})$ and $\psi_i$ for $\mathcal{D}_i$'s optimizable parameters $\{\phi,\mathbf{h}_i,\boldsymbol{\theta}_i\}$.

\subsection{Differentiable Non-rigid Ray-casting}
\label{sec:nonrigid_raycast}
For rigid scenes, the sphere tracing algorithm~\cite{hart1996sphere,jiang2020sdfdiff,yariv2020multiview} is widely used to find the intersection point of a ray and the SDF. However, it is not feasible here due to the deformation fields. Inspired by the method in~\cite{seyb2019non}, which proposes a strategy to render a deformed SDF, we utilize the explicit mesh to help find the intersection point of a ray and $\mathcal{S}_i$.

\begin{figure}
	\begin{center}
		\includegraphics[width=\linewidth]{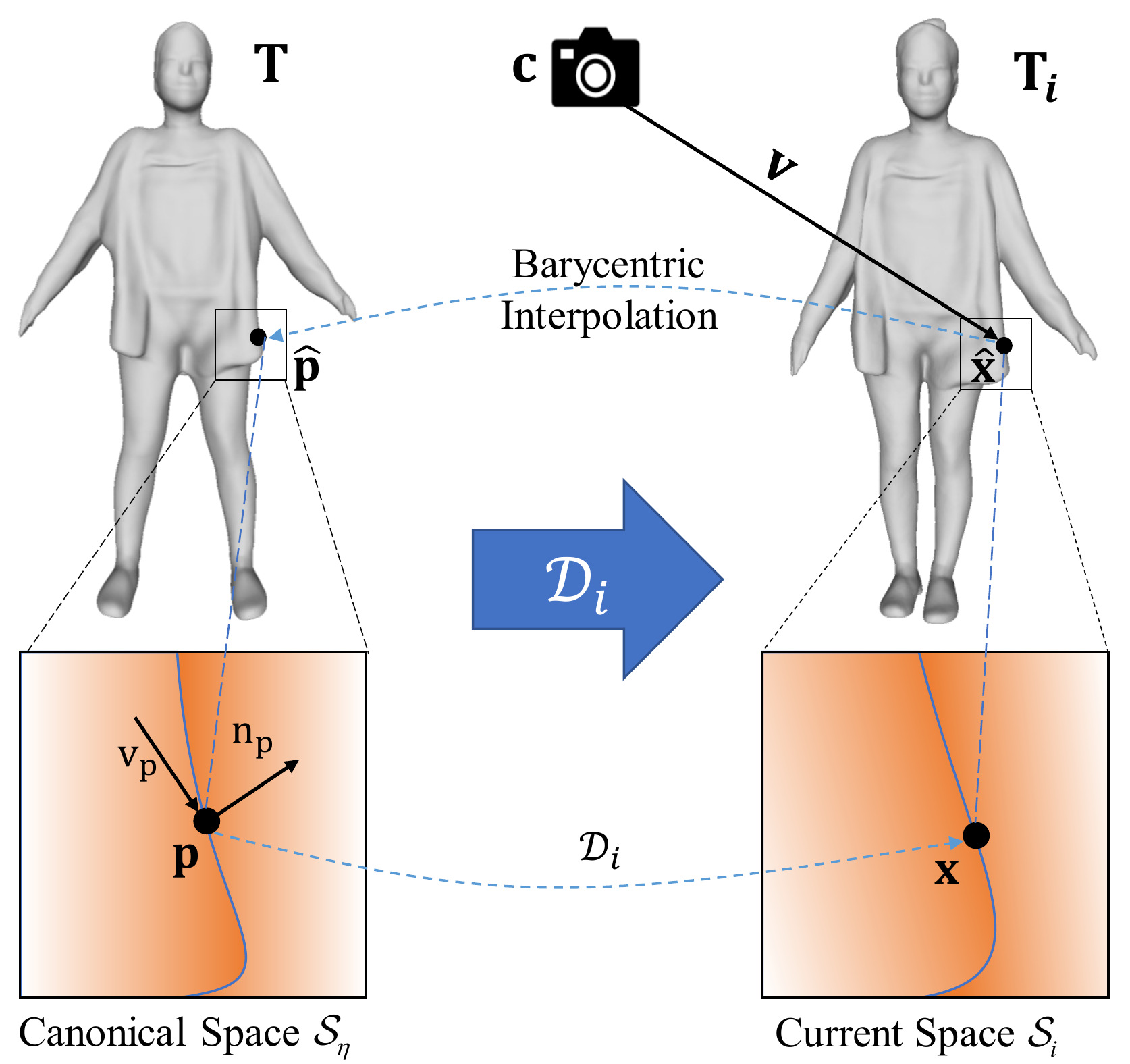}
	\end{center}
	\caption{Some related symbols and illustration about differentiable non-rigid ray-casting and implicit neural rendering.}
	\label{fig:nonrigid_raycast}
\end{figure}

As shown in Fig~\ref{fig:nonrigid_raycast}, we extract an explicit template mesh $\mathbf{T}$ from the canonical surface $\mathcal{S}_{\eta}$. With deformation $\mathcal{D}_i$, we can get $i$-th frame's mesh $\mathbf{T}_i$. Theoretically, $\mathbf{T}_i$ is a piecewise linear approximation of $\mathcal{S}_i$. Therefore, consider a ray emitted from the camera position $\mathbf{c}$ along the direction $\mathbf{v}$, its first intersection $\hat{\mathbf{x}}$ with $\mathbf{T}_i$ is a good approximation of its intersection with $\mathcal{S}_i$. Moreover, with the intersected triangle on $\mathbf{T}_i$, we can find $\hat{\mathbf{x}}$'s corresponding point $\hat{\mathbf{p}}$ on the template $\mathbf{T}$ by consistent barycentric weights. Obviously, $\hat{\mathbf{p}}$ is close to $\mathcal{S}_{\eta}$ and is a good approximation of $\mathcal{D}_i^{-1}(\hat{\mathbf{x}})$. With $\hat{\mathbf{p}}$ as good initialization, we can find a point $\mathbf{p}$ on $\mathcal{S}_{\eta}$, whose deformed point $\mathbf{x}=\mathcal{D}_i(\mathbf{p})$ is exactly the intersection point of the ray $\mathbf{r}$ and $\mathcal{S}_i$. Specifically, we solve $\mathbf{p}$ by:
\begin{equation}
\label{equ:nonrigid_raycast}
\mathbf{p}=\mathop{\arg\min}_{\hat{\mathbf{p}}} \ \omega|f(\hat{\mathbf{p}})|+ \frac{\|(\mathcal{D}_i(\hat{\mathbf{p}})-\mathbf{c})\times \mathbf{v}\|_2}{\|\mathcal{D}_i(\hat{\mathbf{p}})-\mathbf{c}\|_2},
\end{equation} 
where the first item constrains $\hat{\mathbf{p}}$ to be close with $\mathcal{S}_{\eta}$ and the second item restricts $\mathcal{D}_i(\hat{\mathbf{p}})$ on the ray. In our implementation, we set $\omega = 3.05$ and execute $10$ gradient descent iterations to solve $\mathbf{p}$. To guarantee accuracy, we reject those samples with large losses.

\textbf{Differentiable Formula.} The above-mentioned solving process of $\mathbf{p}$ is an iterative optimization process, which is not differentiable. For the ray in $i$-th frame, camera position $\mathbf{c}$, view direction $\mathbf{v}$, $\mathcal{D}_i$'s parameters $\psi_i$ and $f$'s parameter $\eta$ uniquely determine the $\mathbf{p}$. Therefore, $\mathbf{p}$ can be seen as a function of these parameters, and we need to compute partial derivatives of $\mathbf{p}$ to all these parameters. For brevity, we only clarify our calculation for $\eta$ here, and other partial derivatives are computed similarly.

Through the above analysis, $\mathbf{p}$ satisfies the surface and ray constraints: $f(\mathbf{p})\equiv 0$ and $(\mathcal{D}_i(\mathbf{p})-\mathbf{c})\times \mathbf{v}\equiv 0$. We differentiate these two equations w.r.t $\eta$ to get:
\begin{equation}
{\frac{\partial f}{\partial \mathbf{p}}}^T\frac{\partial \mathbf{p}}{\partial \eta} = -\frac{\partial f}{\partial \eta} \quad
[\mathbf{v}]_{\times}\frac{\partial \mathbf{x}}{\partial \mathbf{p}} \frac{\partial \mathbf{p}}{\partial \eta} = 0,
\end{equation}
where $\mathbf{x} = \mathcal{D}_i(\mathbf{p})$ and $[\mathbf{v}]_{\times}$ is $\mathbf{v}$'s cross product matrix. We concatenate these two equations to get a $4\times 3$ linear system, then $\frac{\partial \mathbf{p}}{\partial \eta}$ is computed by solving its normal equation.

\subsection{Implicit Rendering Network}
\label{sec:implicit_render}
IDR~\cite{yariv2020multiview} proposes an MLP $M$ to approximate the rendering equation and demonstrates certain disentangle ability of lighting and material. In their rigid configuration, $M$ takes the zero isosurface's point, its normal, the view direction and its global geometry feature vector as input to estimate the point's color along the view direction. We subtly transfer their design to non-rigid scenarios by converting related current frame's attributes to canonical space. As shown in Fig.~\ref{fig:nonrigid_raycast}, considering a ray emitted from camera center $\mathbf{c}$ along a sampled pixel, whose direction $\mathbf{v}$ is determined by camera's intrinsic parameters $\tau$, we compute its intersection point $\mathbf{p}$ on $\mathcal{S}_{\eta}$ with the algorithm described in sec~\ref{sec:nonrigid_raycast}. In the meantime, we compute its normal $\mathbf{n}_{\mathbf{p}} = \nabla f(\mathbf{p};\eta)$ by gradient calculation. Then, the view direction $\mathbf{v}_{\mathbf{p}}$ in canonical space can be computed by transferring $\mathbf{v}$ with the Jacobian matrix $J_{\mathbf{x}}(\mathbf{p})$ of the deformed point $\mathbf{x}=\mathcal{D}_i(\mathbf{p})$ w.r.t $\mathbf{p}$.
As for the global geometry feature, we similarly use a larger MLP $F(\mathbf{p};\eta) = (f(\mathbf{p};\eta),\mathbf{z}(\mathbf{p};\eta))$ to additionally compute it, which implies the geometry information around $\mathbf{p}$ and can be used to help the prediction of global shadow~\cite{yariv2020multiview}. Finally, we use an MLP $M$ with learnable weights $\gamma$ to compute $\mathbf{p}$'s color $L_{\mathbf{p}}(\eta,\psi_i,\gamma,\tau)$, formulated as:
\begin{equation}
	\label{equ:neural_render}
	\begin{aligned}
		L_{\mathbf{p}}(\eta,\psi_i,\gamma,\tau)&=M(\mathbf{p},\mathbf{n}_{\mathbf{p}},\mathbf{v}_{\mathbf{p}},\mathbf{z}(\mathbf{p};\eta);\gamma) \\
		\mathbf{n}_{\mathbf{p}}&=\nabla f(\mathbf{p};\eta) \\
		\mathbf{x}&=\mathcal{D}_i(\mathbf{p}) \\
		\mathbf{v}_{\mathbf{p}}&=J_{\mathbf{x}}(\mathbf{p})^{-1}\mathbf{v},
	\end{aligned}
\end{equation}
where related symbols have been described above. It can be seen that the color along direction $\mathbf{v}$ of the deformed point $\mathbf{x}$ in $i$-th frame is determined by the MLP weights $\eta$ and $\gamma$, camera parameters $\tau$ and deformation field parameters $\psi_i$. 

\subsection{Loss Function}
According to the description above, for a $N$-frame self-rotating video, the set $\mathcal{X}$ of all optimizable parameters is:
$$
\mathcal{X}=\{\eta,\gamma,\phi,\tau\}\cup\{\mathbf{h}_i,\boldsymbol{\theta}_i|i\in{1,...,N}\},
$$
which includes camera parameters, the learnable weights of MLPs shared by the whole sequence and per frame's specific pose parameters and non-rigid deformation field's conditional variable. Our target is to design a loss function and optimize $\mathcal{X}$ to match the mask and RGB images $\{O_i,I_i|i\in{1,...,N}\}$ of the input video. Besides, a predicted normal map $\{N_i|i\in{1,...,N}\}$ is added to the optimization. As SelfRecon maintains both explicit and implicit geometry, the loss terms can be divided into two parts.


\subsubsection{Explicit Loss}
During the computation of explicit losses, we temporarily regard the canonical mesh vertices $\mathbf{T}$ as an optimizable variable and compute its gradient together with $\mathcal{X}$. Then in the consistency loss, we associate its variations with our implicit representation. At present, explicit losses mainly include mask loss, deformation regularization loss, and the smoothness loss of the skeleton.

\textbf{Mask Loss.} We utilize a differentiable renderer~\cite{wiles2020synsin} based on point cloud to render the mask $O(\mathbf{T}_i)$ of $i$-th frame's mesh $\mathbf{T}_i = \mathcal{D}_i(\mathbf{T})$ with camera parameters, and compute the IoU loss with target mask $O_i$:
\begin{equation}
	loss_{IoU} = 1 - \frac{\|O(\mathbf{T}_i)\otimes O_i\|_1}{\|O(\mathbf{T}_i)\oplus O_i -  O(\mathbf{T}_i)\otimes O_i\|_1},
\end{equation}
where $\otimes$ and $\oplus$ are the operators that perform element-wise product and sum respectively.

\textbf{Deformation Regularization Loss.} As stated in Sec. \ref{sec:deform}, the $i$-th frame's deformation field $\mathcal{D}_i$ contains variable $d$ and fixed $\mathcal{W}$. $d$ represents the deformation that can not be represented by skinning transformation $\mathcal{W}$, and this deformation should be relatively small. To associate the skeleton pose, we design the following regularization loss:
\begin{equation}
	\label{equ:w3_defregu}
	\begin{aligned}
		loss_{regu}=\frac{1}{|\mathbf{T}|}\sum_{\mathbf{t}\in{\mathbf{T}}}\rho(\|\mathcal{W}(\mathbf{t};\boldsymbol{\theta}_i)-\mathcal{D}_i(\mathbf{t})\|_2), 
	\end{aligned}
\end{equation}
where $\mathbf{t}$ is a vertex coordinate of $\mathbf{T}$, $|\mathbf{T}|$ is the vertices number of $\mathbf{T}$, and $\rho$ is the Geman-McClure robust loss~\cite{ganan1985bayesian}.

\textbf{Skeleton Smoothness Loss.} The motion trajectory of the joints should be low frequency. Similar with MonoPerf~\cite{xu2018monoperfcap}, we smooth the skeleton coordinates of $30$ consecutive frames by minimizing the distance to a $10$ dimensional linear subspace $\mathbf{B}\in{\mathbb{R}^{30\times10}}$ spanned by the $10$ lowest frequency basis vectors of the discrete cosine transform:
\begin{equation}
	loss_{ske} = \frac{1}{30} \|\mathbf{J}\text{Null}(\mathbf{B})\|_F^2.
\end{equation}
Here, $\text{Null}(\mathbf{B})$ denotes the nullspace of the $\mathbf{B}$ matrix, the matrix $\mathbf{J}\in{\mathbb{R}^{72\times30}}$ stacks all skeleton coordinates of consecutive $30$ frames, and $\|\cdot\|_F$ denotes the Frobenius norm.

Finally, the loss for the explicit representation is:
\begin{equation}
	Loss_{exp}=loss_{IoU}+\lambda_{e1}loss_{regu}+\lambda_{e2}loss_{ske}.
\end{equation}
$\lambda_{e1}$ and $\lambda_{e2}$ adjust the weights of related losses. After each iteration, we reserve $\mathcal{X}$'s gradients and wait for the implicit loss iteration to update together. For canonical mesh vertices, we use SGD to update $\mathbf{T}$ to $\hat{\mathbf{T}}$, which will be used in consistency loss to match two representations.

\subsubsection{Implicit Loss}
We sample pixels within the ground truth mask and utilize Sec.~\ref{sec:nonrigid_raycast} to get the ray's intersection $\mathbf{p}$ on $\mathcal{S}_{\eta}$ and its corresponding ground truth color $I_{\mathbf{p}}$ and predicted normal $N_{\mathbf{p}}$ if available. Then, based on this sampled points set $\text{P}$, we construct two losses.

\textbf{Color Loss.} By referring to Eq.~\eqref{equ:neural_render}, we formulate the color loss as:
\begin{equation}
	loss_{RGB} = \frac{1}{|\text{P}|}\sum_{\mathbf{p}\in \text{P}}|L_{\mathbf{p}}(\mathcal{X})-I_{\mathbf{p}}|.
\end{equation}
Here, we use $\mathcal{X}$ to substitute related parameters in Eq.~\eqref{equ:neural_render}. Intuitively, this loss requires that the rendered images should match the input images.

\textbf{Normal Loss.} We utilized the predicted normal map by PIFuHD~\cite{saito2020pifuhd} to further refine the geometry shape. Referring to Eq.~\eqref{equ:neural_render}, we can easily compute $\mathbf{p}$'s normal $\mathbf{n}_\mathbf{p}$. Besides, we need to transform the corresponding predicted normal $N_{\mathbf{p}}$ from the space of current frame to canonical space, which can be computed with $J_{\mathbf{x}}(\mathbf{p})^T N_{\mathbf{p}}$, where $J_{\mathbf{x}}(\mathbf{p})$ is the Jacobian matrix of the forward deformation field at $\mathbf{p}$~\cite{seyb2019non}. Thus, the normal loss is:
\begin{equation}
	\label{equ:norm}
	loss_{norm}=\frac{1}{|\text{P}|}\sum_{\mathbf{p}\in{\text{P}}}\omega_\mathbf{p}\|\mathbf{n}_\mathbf{p}-unit(J_{\mathbf{x}}(\mathbf{p})^T N_{\mathbf{p}})\|_2.
\end{equation}
Here, $unit(\cdot)$ means to normalize the vector. $\omega_\mathbf{p}$ is the weight defined by the cosine of angle between $\mathbf{n}_\mathbf{p}$ and corresponding view direction. Since the predicted normals are noisy and inconsistent between frames, we use this weights to alleviate the impact of normal that deviates from its view direction and avoid geometry artifacts.

We also design regular losses for the implicit representation, and these losses are defined on the set of sampled points $\text{S}$ near the implicit surface~\cite{gropp2020implicit}.

\textbf{Rigidity Loss.} We require the first deformation field $d$ to be as rigid as possible to avoid distortion. Following Park \textit{et al.}~\cite{park2021nerfies}, we design our loss as:
\begin{equation}
	\begin{aligned}
		loss_{rigid}&=\frac{1}{|\text{S}|}\sum_{\mathbf{p}\in \text{S}}\rho(\|\text{log}\Sigma_{\mathbf{p}}\|_F),
	\end{aligned}
\end{equation}
where $\Sigma_{\mathbf{p}}$ is the singular value diagonal matrix of the Jacobian of $d$ on $\mathbf{p}$ and $\rho$ is the robust function~\cite{ganan1985bayesian}.

\textbf{Eikonal Loss.} We adopt the regular loss of IGR~\cite{gropp2020implicit} to make $f$ to be sign distance function:
\begin{equation}
	loss_{sdf} = \frac{1}{|\text{S}|}\sum_{\mathbf{p}\in{\text{S}}}(\|\mathbf{n}_{\mathbf{p}}\|_2-1)^2,
\end{equation}
where $\mathbf{n}_\mathbf{p}$ is obtained by differentiating $f$ at $\mathbf{p}$.

Finally, the implicit loss can be represented as:
\begin{equation}
	\label{equ:implicit}
	Loss_{imp} = loss_{RGB}+\lambda_{i1}loss_{norm}+\lambda_{i2}loss_{rigid}+\lambda_{i3}loss_{sdf},
\end{equation}
where $\lambda_{i1}$, $\lambda_{i2}$ and $\lambda_{i3}$ are balancing weights.

\subsubsection{Explicit/Implicit Consistency}
After explicit iteration, the canonical mesh has been updated to $\hat{\mathbf{T}}$, to make the implicit SDF consistent with the updated explicit mesh during implicit iteration, we design a consistency loss:
\begin{equation}
	Loss_{cons} = \frac{1}{|\hat{\mathbf{T}}|}\sum_{\hat{\mathbf{t}}\in{\hat{\mathbf{T}}}}|f(\hat{\mathbf{t}};\eta)|,
\end{equation}
where $\hat{\mathbf{t}}$ is a vertex coordinate of $\hat{\mathbf{T}}$. Intuitively, the loss requires $\hat{\mathbf{T}}$ to match the implicit surface $\mathcal{S}_{\eta}$.

In each optimization step, we first perform the explicit iteration to obtain $\hat{\mathbf{T}}$ and reserve $\mathcal{X}$'s gradients. Then, we compute implicit and consistency losses to accumulate new $\mathcal{X}$'s gradients. Finally, Adam is utilized to update $\mathcal{X}$ with computed gradients.

%% file: experiment.tex
\section{Experiments}

\begin{figure*}
	\begin{center}
		\includegraphics[width=\linewidth]{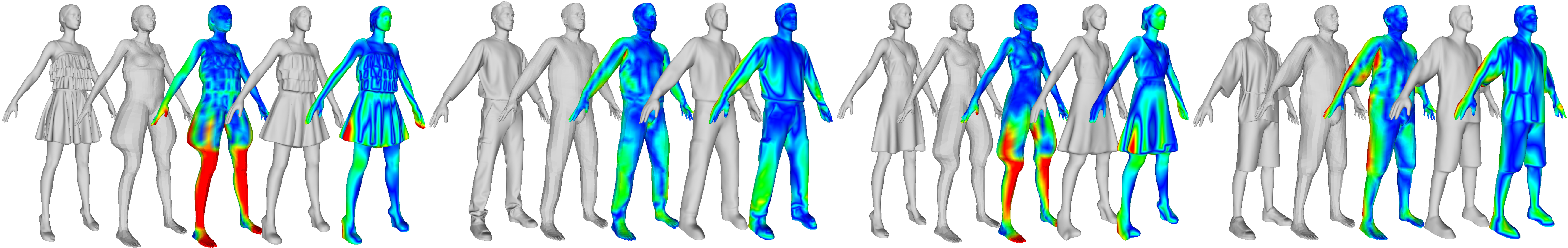}
	\end{center}
	\caption{Reconstructions in canonical pose and their error maps for four synthetic self-rotating sequences. In each group, we show the GT mesh, results of VideoAvatar and SelfRecon in turn (red means $\geq6$cm).}
	\label{fig:syn_compare}
\end{figure*}

We conduct quantitative and qualitative experiments to demonstrate the effectiveness of SelfRecon. For quantitative evaluation, we synthesize several sequences with commercial software due to the lack of high-quality geometry data of human body with general clothing. For qualitative evaluation, we mainly utilize the PeopleSnapshot~\cite{alldieck2018video} dataset and several real sequences collected by ourselves. We also present ablation study for the loss term design and present an avatar generation application. 

\begin{table}
	\centering
	\caption{The errors (cm) on the synthetic five sequences. We report three error metrics: the average distance from reconstructed to GT meshes (Recon),  the average distance from GT to reconstructed meshes (GT), and the Chamfer distance. For each error metric, we report the values of VideoAvatar and ours in two consecutive rows.}
	\label{tab:syn_compare}
	\begin{tabular}{|c|c|c|c|c|c|c|}
		\hline
		Subject & f1 & f2 & f3 & m1 & m2 & mean\\
		\hline
		\multirow{2}{*}{Recon} & \textbf{1.59} & 1.71 & 1.93 & 1.81 & 1.27 & 1.66 \\ \cline{2-7}
		& 1.67 & \textbf{1.32} & \textbf{1.63} & \textbf{1.53} & \textbf{1.17} & \textbf{1.46} \\
		\hline
		\multirow{2}{*}{GT} & 2.08 & 1.50 & 2.40 & 1.92 & 1.42 & 1.86 \\ \cline{2-7}
		& \textbf{1.62} & \textbf{1.16} & \textbf{1.92} & \textbf{1.53} & \textbf{1.17} & \textbf{1.48} \\
		\hline
		\multirow{2}{*}{Chamfer} & 1.84 & 1.60 & 2.17 & 1.86 & 1.34 & 1.76 \\ \cline{2-7}
		& \textbf{1.64} & \textbf{1.24} & \textbf{1.77} & \textbf{1.53} & \textbf{1.17} & \textbf{1.47} \\
		\hline
	\end{tabular}
\end{table}

\subsection{Quantitative Evaluation}
We synthesize data to quantitatively evaluate our reconstruction algorithm. Specifically, we use Blender~\cite{Blender} to design the self-rotating motions for male and female avatars. Then, we utilize CLO3D~\cite{CLO3D} to design several clothes and animate the clothed body with motions. Finally, we synthesized two sets of male and three sets of female dressing sequences. We reconstruct these sequences with VideoAvatar~\cite{alldieck2018video} and our method, and report the registration error for the canonical posture results in Tab.~\ref{tab:syn_compare}. Compared with VideoAvatar, our method significantly reduces the values of various error metrics. In Fig.~\ref{fig:syn_compare}, we also present four group results and their error maps. Intuitively, our results capture the overall shape and have some reasonable details. As VideoAvatar is based on the SMPL+D representation, it has plausible results for tighter clothing, like the male examples, but lacks detailed reconstruction ability. Moreover, it can not correctly reconstruct loose clothing, especially for the females dressed in skirts.

\begin{figure}
	\begin{center}
		\includegraphics[width=\linewidth]{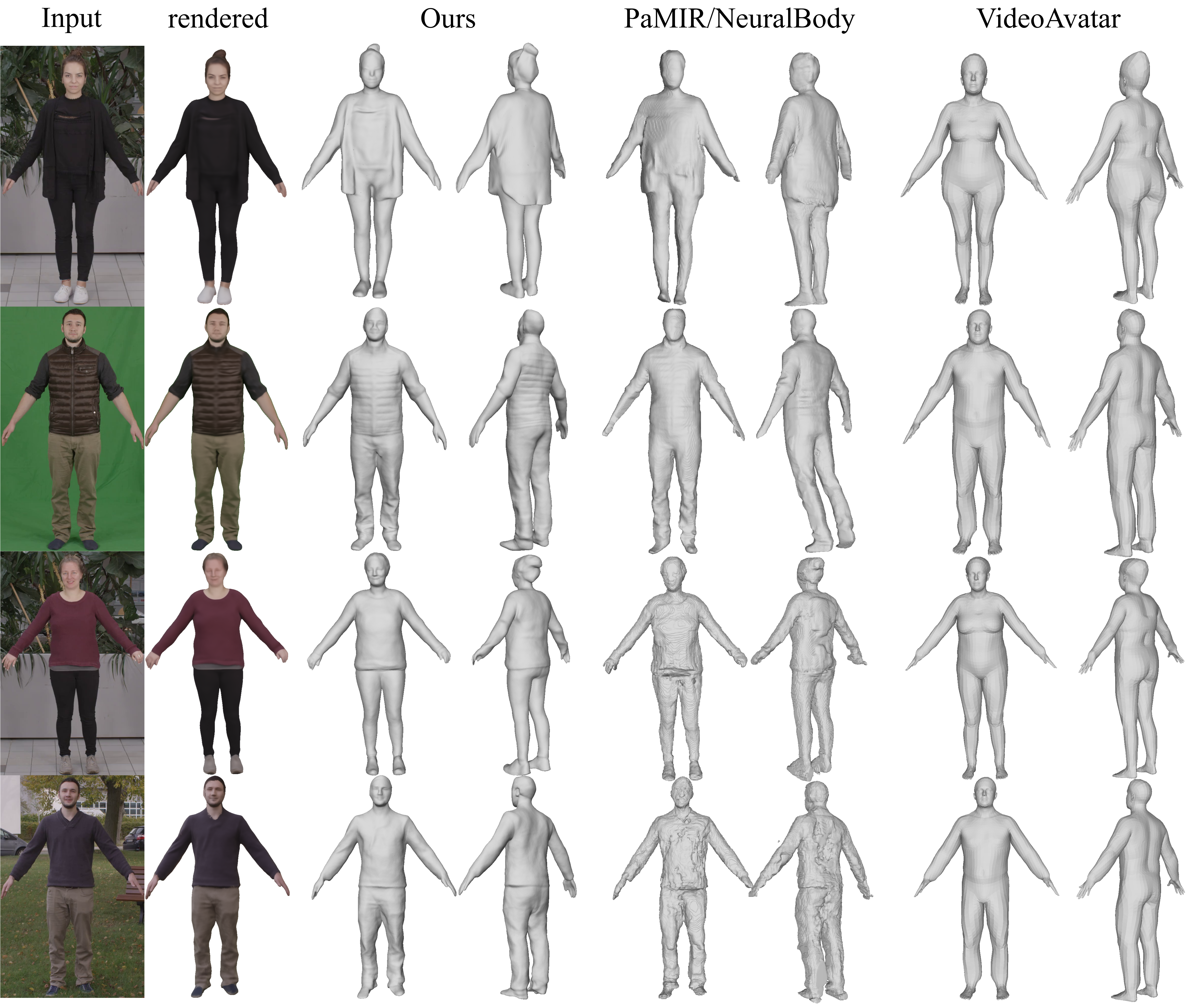}
	\end{center}
	\caption{Comparison results with methods that use video or multi-frame images, including PaMIR~\cite{zheng2021pamir}, NeuralBody~\cite{peng2021neural} and VideoAvatar~\cite{alldieck2018video}. For the second comparison group, the method in the first two rows is PaMIR, and the rest is NeuralBody. We also present our rendered image as reference. SelfRecon can reconstruct high-fidelity geometry shape of standing posture, including facial features and clothing folds.}
	\label{fig:compare}
\end{figure}

\subsection{Qualitative Evaluation} 
We also qualitatively compare SelfRecon with multi-frame prediction algorithm PaMIR~\cite{zheng2021pamir}, optimization method VideoAvatar~\cite{alldieck2018video} and NeRF~\cite{mildenhall2020nerf} based neural rendering method NeuralBody~\cite{peng2021neural} on several sequences of PeopleSnapshot dataset. In Fig.~\ref{fig:compare}, we present the first frame of input video, our rendered image, and reconstruction results of all methods from two perspectives. And we compare with PaMIR in the first two rows and the others with NeuralBody. As we can see, VideoAvatar based on SMPL+D can only approximately capture the overall shape, but details such as hairstyle and clothing folds are lost. PaMIR uses multi-frame input to improve its results, but still suffers from the deep ambiguity. As in the second example, its reconstructed human is not upright from the side view. Besides, its results have some details but miss facial features, while our method has better details and can recover certain facial features. Similar with SelfRecon, NeuralBody also inputs a video to do self-supervised optimization. It mainly focuses on novel view synthesis, but still can extract geometry from underlying NeRF representation. We can see that its reconstructions alleviate deep ambiguity and conform to the human's overall structure while suffering from large noise on the surface, which may be caused by excessive freedom of volume rendering. Different from their method, implicit surface representation based SelfRecon can recover high-fidelity geometry shape without noises.

\begin{figure}
	\begin{center}
		\includegraphics[width=\linewidth]{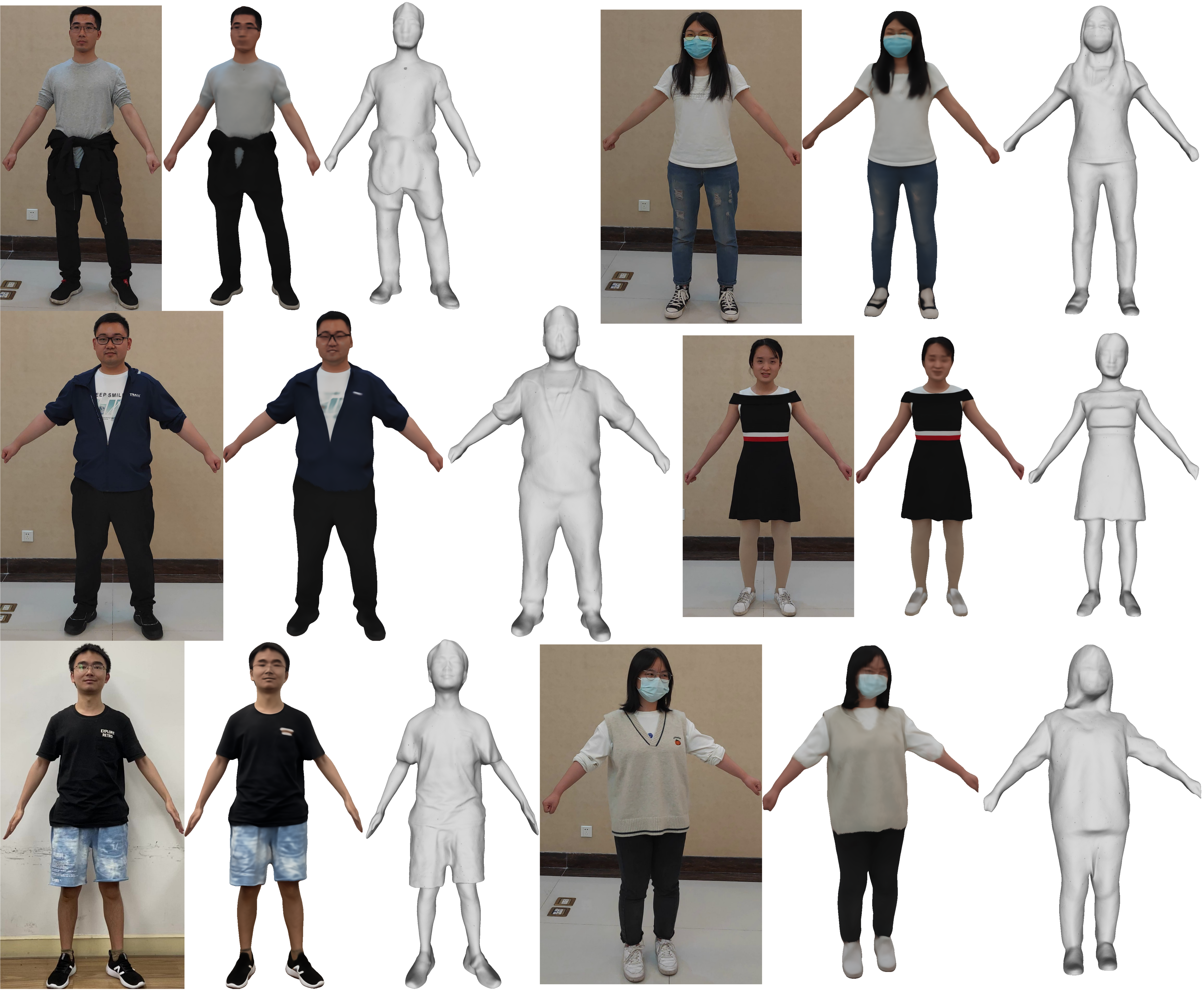}
	\end{center}
	\caption{Reconstruction results from videos taken by smartphones. Each group shows the first image of the video, corresponding neural rendered image and reconstructed shape.}
	\label{fig:present}
\end{figure}

Fig.~\ref{fig:present} shows reconstruction results on our collected videos with smartphones. For each group, we present the first frame of the video, our rendered image and reconstruction. Our results have high-fidelity geometry shapes for kinds of clothing and body, and our neural rendered images are also quite close to the input images.

\begin{figure}
	\begin{center}
		\includegraphics[width=\linewidth]{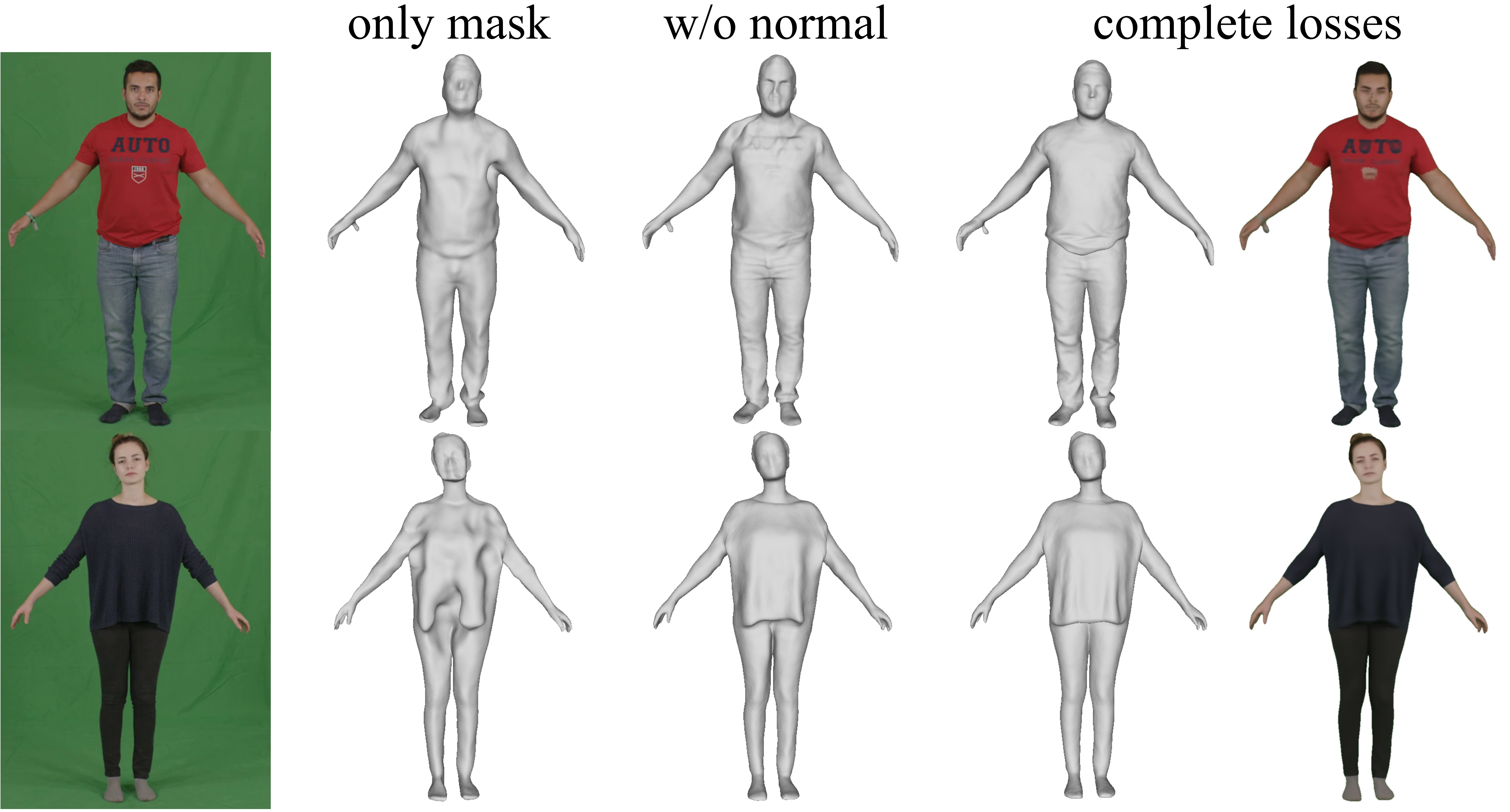}
	\end{center}
	\caption{Ablation study for color, mask and normal losses. With mask loss only, the results lack details and have much concave geometry. After adding color loss, geometry is significantly improved but can not completely eliminate the concavities. With the predicted normals, the results are further improved. In the last column, we also show the neural rendering images as reference.}
	\label{fig:ablation}
\end{figure}

\subsection{Ablation Study}
Our complete algorithm requires color images, masks and normal maps as inputs.  Fig.~\ref{fig:ablation} shows ablation experiments of two examples on three inputs. As the results show, if only using the mask loss, the recovered geometry shape is inside the convex hull formed by the silhouettes but lacks details and has noticeable concavities. After adding the color loss, it significantly improves the details and reduces the unnatural concavities. For the second example, the result has been very close to the result obtained by adding the normal loss. However, for the first example, it can not completely eliminate the depressed geometry without normal loss. This may be caused by lacking rich texture and multi-view observations in these areas. With the normal loss, our results are further improved, and the unnatural pits are eliminated while the details are preserved.

\begin{figure}
	\begin{center}
		\includegraphics[width=\linewidth]{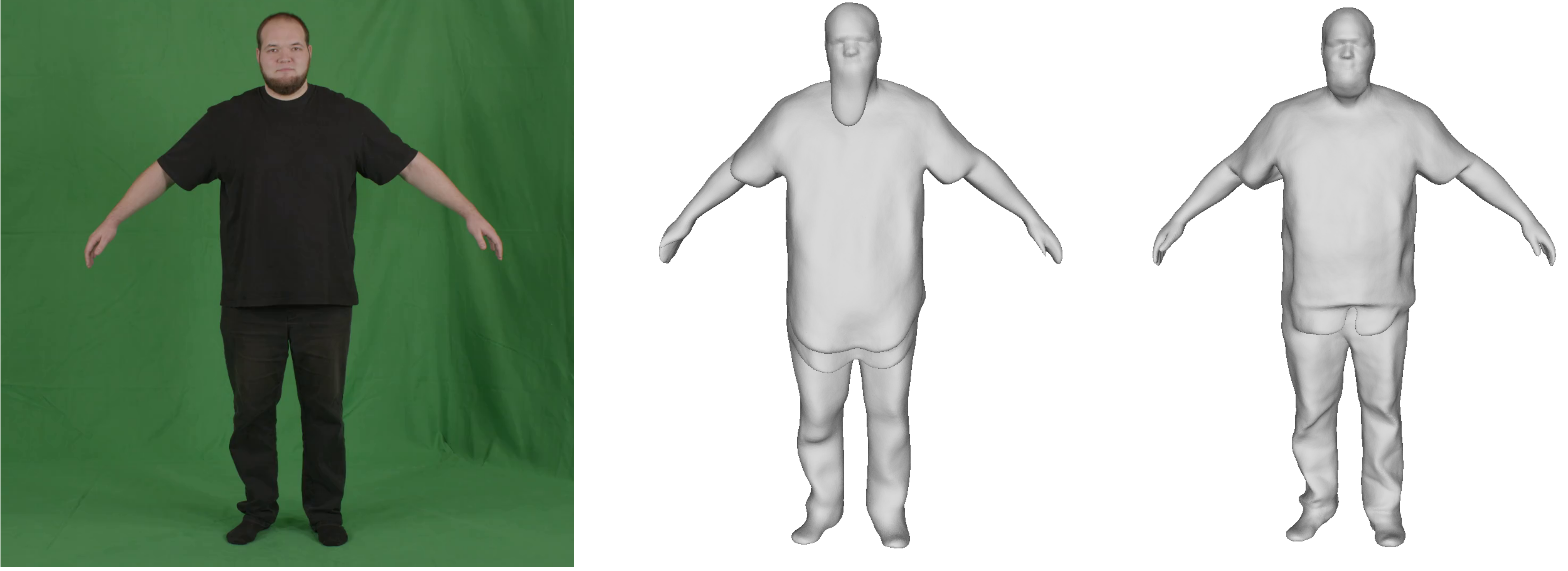}
	\end{center}
	\caption{Without the adaptive weights in Eq.~\eqref{equ:norm} and small $\lambda_{i1}$ in Eq.~\eqref{equ:implicit}, normal loss will result artifacts (middle). After adjusting the weights, the corresponding result (right) is more plausible.}
	\label{fig:norm}
\end{figure}

Since the normal prediction network~\cite{saito2020pifuhd} is trained with synthetic images, its prediction may be not accurate for real tests and may be not consistent for different frames. As shown in Fig.~\ref{fig:norm}, without the adaptive weights in Eq.~\eqref{equ:norm}, the normal loss might result unexpted results.

\subsection{Avatar Generation}

\begin{figure}
	\begin{center}
		\includegraphics[width=\linewidth]{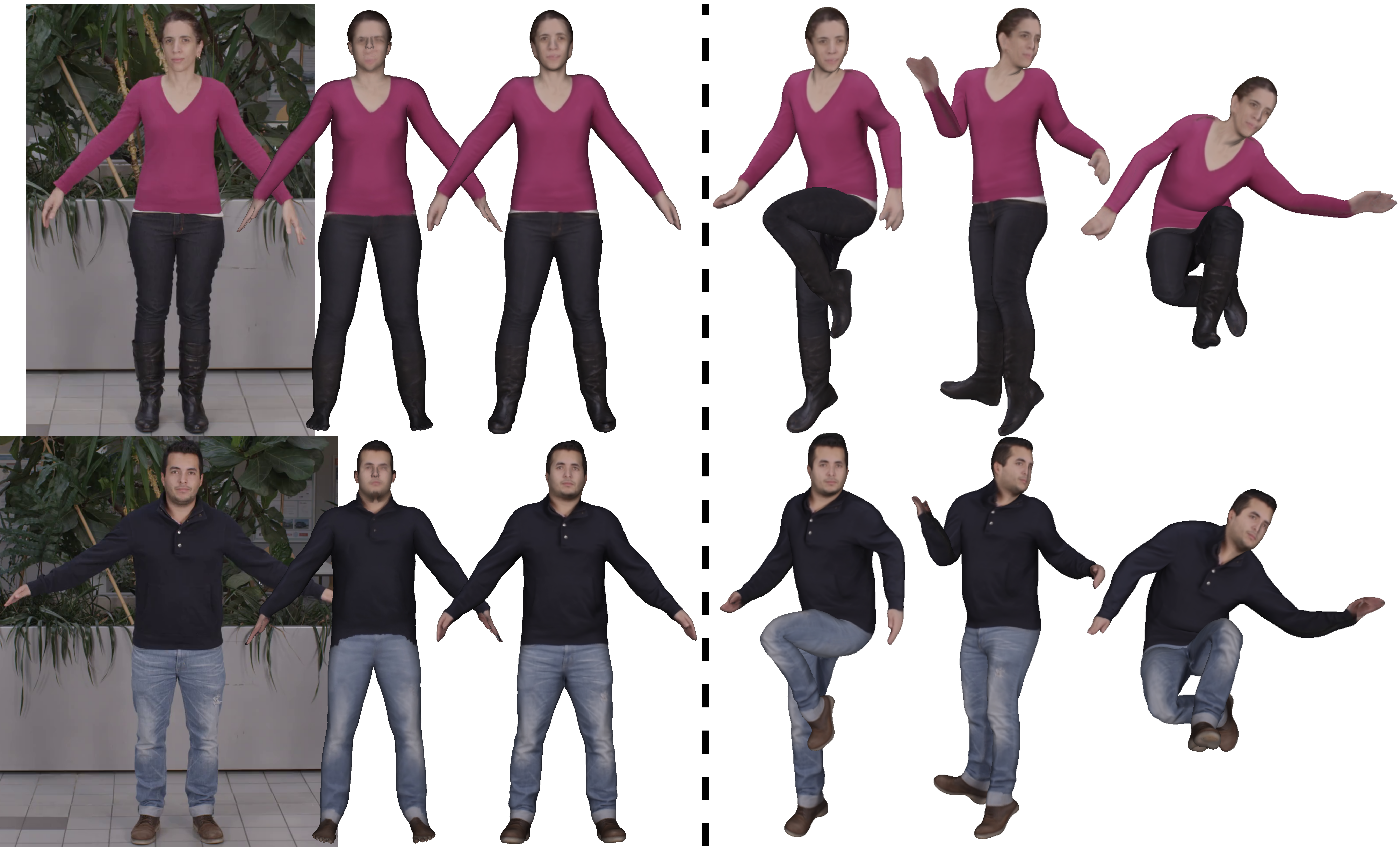}
	\end{center}
	\caption{The reconstructed texture mesh and driving results. The left shows the reference image, texture meshes generated by VideoAvatar and SelfRecon. On the right, we use three pose parameters to drive our texture mesh and generate plausible results.}
	\label{fig:animation}
\end{figure}

Thanks to our forward deformation field design, we can extract a mesh sequence with consistent topology. Based on the tracking results, we can extract a texture template mesh from the images with bound skinning weights from the skinning transformation field. Then, an animatable avatar is generated and can be driven with SMPL pose parameters. For texture extraction, we follow the method of VideoAvatar~\cite{alldieck2018video}. Fig.~\ref{fig:animation} shows two examples of texture generation and driving from the PeopleSnapshot dataset. Our method recovers better geometric details like facial, shoes and clothing folds thanks to more accurate tracking results. Besides, our driving results look plausible and may be of sufficient quality for some applications.

%% file: conclusion.tex
\section{Conclusion and Discussion}
We proposed SelfRecon, a self-supervised reconstruction method based on neural implicit representation and neural rendering. With forward deformation, our method can be easily applied to body movement and recover space-time coherent surfaces, which is convenient for downstream applications. Moreover, combining the explicit representation, we proposed a non-rigid ray casting algorithm, which makes it possible for differentiable intersecting with the deformed implicit surface. SelfRecon can reconstruct high-fidelity clothed body shape from a self-rotating video without pre-computed templates. We also show high-fidelity avatar generation with our tracking results, demonstrating potential applications of SelfRecon.

SelfRecon still has several limitations. First, it requires relatively long time to optimize, which limits its convenient applications. However, this problem can be alleviated with the help of body priors and the fast growing field of neural rendering. Second, current method relies on the predicted normal maps to improve the geometric details. How to recover the geometric details directly from the self-supervised rendering loss is worthy of future study. Third, the proposed method mainly works well for self-rotating motions, and it is worthy of study for more general motion sequences.

{\small{\paragraph{Acknowledgement} This research was supported by National Natural Science Foundation of China (No. 62122071), the Youth Innovation Promotion Association CAS (No. 2018495), ``the Fundamental Research Funds for the Central Universities''(No. WK3470000021).}}